\documentclass{nature}

\usepackage{amssymb}
\usepackage{graphicx}
\usepackage{times}
\usepackage{epsfig}
\usepackage{amsmath}
\usepackage{subfigure}

\usepackage{mathtools}
\usepackage{epstopdf}
\usepackage[font=scriptsize]{caption}
\usepackage{float}
\usepackage[]{algorithm2e}

\newcommand{\argmin}{\arg\!\min}

\title{A Generic Regression Framework for Pose Recognition on Color and Depth Images}

\author{Wenye He}

\begin{document}

\maketitle

\begin{abstract}
Cascaded regression method is a fast and accurate method on finding 2D pose of objects in RGB images. It is able to find the accurate pose of objects in an image by a great number of corrections on the good initial guess of the pose of objects. This paper explains the algorithm and shows the result of two experiments carried by the researchers. The presented new method to quickly and accurately predict 3D positions of body joints from
a single depth image, using no temporal information. We take an object recognition approach,
designing an intermediate body parts representation that maps the difficult pose estimation problem
into a simpler per-pixel classification problem. Our large and highly varied training dataset
allows the classifier to estimate body parts invariant to pose, body shape, clothing. Finally, we
generate confidence-scored 3D proposals of several body parts by re-projecting the classification
result and finding local modes. 
\end{abstract}

\section*{Introduction}
Detection and localization provide a helpful function in computer vision. Detection finds out whether an object is contained in the image, while localization tells people which specific place of the image an object is in. For example, there is an image of tree. Detection can tell us whether a bird is in the image and localization shows where the bird it is. Localization could answer whether the bird flies in the sky or stays on the tree. More specifically, localization enable us to know which groups of pixels represent a bird in the image.

On the other hand, robust interactive human body tracking has different applications that include gaming, human computer
interaction, security, telepresence and even the health care. The task has recent been
simplified by the introduction of real time depth cameras. However even the
best existing system has limitations. Until the launch of Kinect, none ran at interactive rates on
consumer hardware while handling a full range of human body shapes and sizes undergoing general
body motions. Detecting body parts from a single depth image is a challenging task from a small set of 3D position candidates for each skeletal joint \cite{ierf4}. They focused
on pre-frame initialization and recovery is designed to complement any appropriate tackling
algorithm

In this paper, we demonstrate an algorithm to answer a question "Is an object $o$ with pose $\theta$ located in the image $I$. In their work, after making a raw guess of the pose of an object in a set of image, they use cascaded pose regression to detect which image has such the object and locate which region in the image contains the object. Pose-indexed features\cite{iref2} and its assumed weak invariance are used in the algorithm. Furthermore, random ferns\cite{iref3} regressors are applied in CPR.  The system runs at 200 frames per second on consumer hardware. The evaluation shows high
accuracy on both synthetic and real test sets, and investigates the effect of several training parameters.
They achieved state of the art accuracy in our comparison with related work and demonstrate
improved generalization over exact whole-skeleton nearest neighbour matching.

\section*{Related Works}
This paper uses features to achieve pose estimate. In computer vision, features have great functions. The popular use of features is recognition. Paul and Michael\cite{iref4} use rectangle features for object detection. Lowe\cite{iref5} present an approach to identify objects using higly distinctive and scale - invariant features, and has been applied for many applications \cite{tan13}. Features also help people in 3D model retrieval. Ryutarou et al.\cite{iref6} obtain 3D model using salient local visual features. In this paper, researchers apply pose-indexed features\cite{iref2} whose concept is proposed by Francois and Donald to estimate pose. Early work in pose estimation contains active contour models\cite{iref7}, 2D range scans matching\cite{iref8}, and active appearance models\cite{iref9}. In recent work, pose estimation is employed to reconstruct 3D shape\cite{iref10} and detect and localize object\cite{iref11}.
 [\cite{iref25}] use the marginal statistics of unlabelled data to improve pose estimation. [\cite{iref41}, \cite{shen13}] proposed a local mixture of Gaussian Processes to regress human pose. Auto-context was used in [\cite{iref40}] to obtain a coarse body part labelling.  [\cite{iref42}] track a hand clothed in a colored glove. Our system could be automatically inferring the colors of a virtual colored suit from a depth image and used for many RGB-D applications, such as virtual reality \cite{Su13}.

\section*{Method}
This section describes how Cascaded Pose Regression works. First of all, the researchers create an image model $G :  \mathcal{O} \times \Theta \to \mathcal{I}$. The model show how an image $I \in \mathcal{I}$ is constructed from an object $o \in \mathcal{O}$ and pose $\theta\in\Theta$. Given that poase estimate is suppose to be unique in the model, $\mathcal{G}(o_1,\theta_1)=\mathcal{G}(o_2,\theta_2)$ iff $o_1=o_2$ and $\theta_1=\theta_2$. Moreover, the operator $\circ$ is designed for combination of two poses. The researchers write a formula for a new pose composed of $\theta_1$ and $\theta_2$, $\theta=\theta_1\circ\theta_1$. $\overline{\theta}$ is set as the inverse of $\overline{\theta}$, and $e$ as the identity element. A function to calculate relative error between two poses is formed, $d : \Theta \times \Theta \to \mathcal{R}$ where $d(\theta_1,\theta_2)$ depends on $\overline{\theta_1}\circ\theta_2$ or equivalently that $d(\theta_\delta\circ\theta_1,\theta_\delta\circ\theta_2)=d(\theta_1,\theta_2)$ for all $\theta_1,\theta_2,\theta_\delta\in\Theta$.\\
Pose-indexed features and weak invariance are introduced in CPR. A pose-indexed features is a function, $h :  \Theta\times\mathcal{I} \to \mathcal{R}$. $h$ is weakly invariant if $$h(\theta,G(o,e))=h(\theta_{\delta}\circ\theta,G(o,\theta_{\delta})),$$ where $\forall\theta,\theta_\delta\in\Theta$, or equivalently, $$h(\theta_1,G(o,\theta_2))=h(\theta_{\delta}\circ\theta_1,G(o,\theta_{\delta}\circ\theta_2)),$$ where $\forall\theta_1,\theta_2,\theta_\delta\in\Theta$. Each pose-indexed feature is composed of control point features and pose. Each control feature, $h_{p_1,p_2}$, is computed as the difference of two pixels, $p_1$ and $p_2$ at predefined image locations, so $h_{p_1,p_2}(I)=I(p_1)-I(p_2)$ where $I(p)$ is the grayscale value of image $I$ at location $p$. \begin{figure*}[h]
\centerline{\epsfig{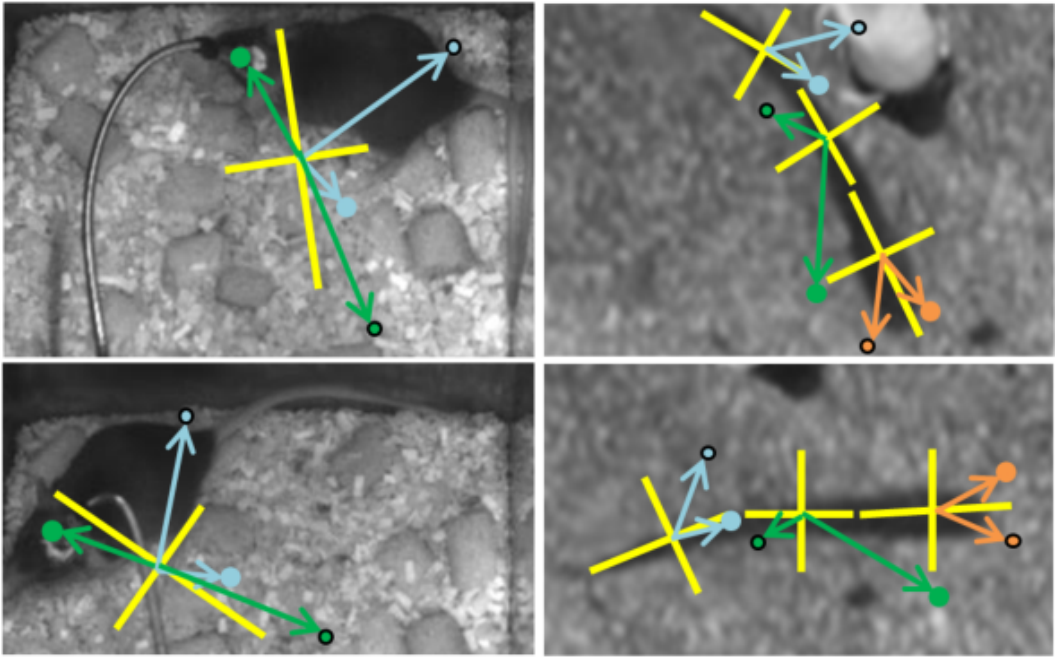}}
\caption{Pose-indexed features. Left: Mice described by a 1-part pose model. Right: zebra fish described by a 3 part pose model. The yellow crosses are the current estimating pose of the object. A pair of arrows in the same color points to control features from their pose coordinate.}
\end{figure*}Therefore, taking pose $\theta$ into account, the researchers define an pose-indexed feature $h_{p_1,p_2}(\theta,I) = I(H_{\theta}p_1)-I(H_{\theta}p_2)$, where $H_\theta$ is an associated $3\times3$ homography matrix. Figure 1 shows pose-indexed features in mice and zebra fish.\\
After introducing pose-indexed features and weakly invariance, the evaluation and training algorithm on CPR is demonstrated here and is shown in Figures 2 and Figure 3.\begin{figure*}[h]
\centerline{\epsfig{figure=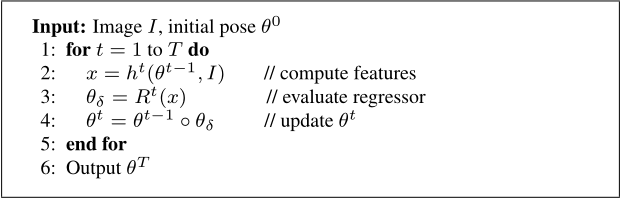,width=10cm}}
\caption{Evaluation of CPR}
\end{figure*}The full formula to evaluate CPR is $\theta^t=\theta^{t-1}\circ R^t(h^t(\theta^{t-1},I))$, where $t=1...T$ and $R$ is a cascaded regressor trainned in Figure 3, given an input pose, initial pose, $\theta^0$ and an image $I$, outputting $\theta^T$. The full evaluating procedures is shown by testing images in Figure 4.\\
The objective for training a new regressor is to reduce the difference between the true pose and the pose calculated by the previous pose-indexed features and previous regressor. The goal is to optimize the loss $\mathcal{L}=\sum^N_{i=1}d(\theta^T_i,\theta_i)$. 
\begin{figure*}[h]
\centerline{\epsfig{figure=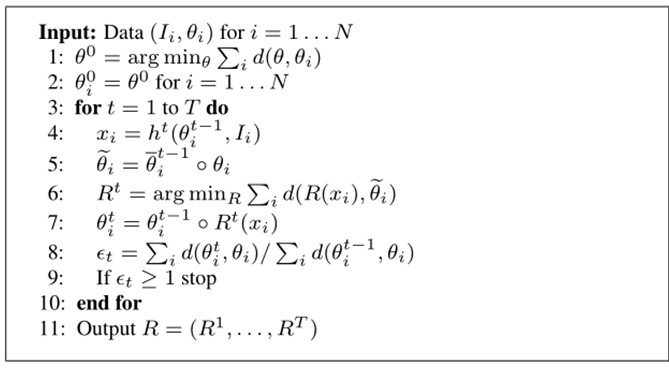,width=10cm}}
\caption{Training for CPR}
\end{figure*}
In step 1 and 2 of figure 3, for each i, there is a pose in the zero phase, $\theta_i^0=\theta^0=\argmin_\theta\sum_i d(\theta,\theta_i)$. $\theta^0$ is the single pose estimate with the minimized training error and without depending on regressors. In every iteration, the researchers compute the pose-indexed features for all the training images set $I$ using the previous pose estimate $\theta_i^{t-1}$ corresponding to the image $I_i$. Then, they compute a new pose $\tilde{\theta}_i$ using the average of the poses in previous one phase, $\overline{\theta}_i^{t-1}$, and the true pose $\theta_i$. In Step 6, the researchers find a regressor $R^t$ by minimizing the loss. In the formula in Step 6, $R$ is a set of regressors, $R=(R^1,...,R^{t-1})$. Using the new regressor, the researchers get the new poses for all the images. In Step 8 and Step 9, the researchers calculate the error by comparing the losses in current phase and the previous one phase. The loop stop once the new iteration cannot decline the training error.\\
\begin{figure*}[h]
\centerline{\epsfig{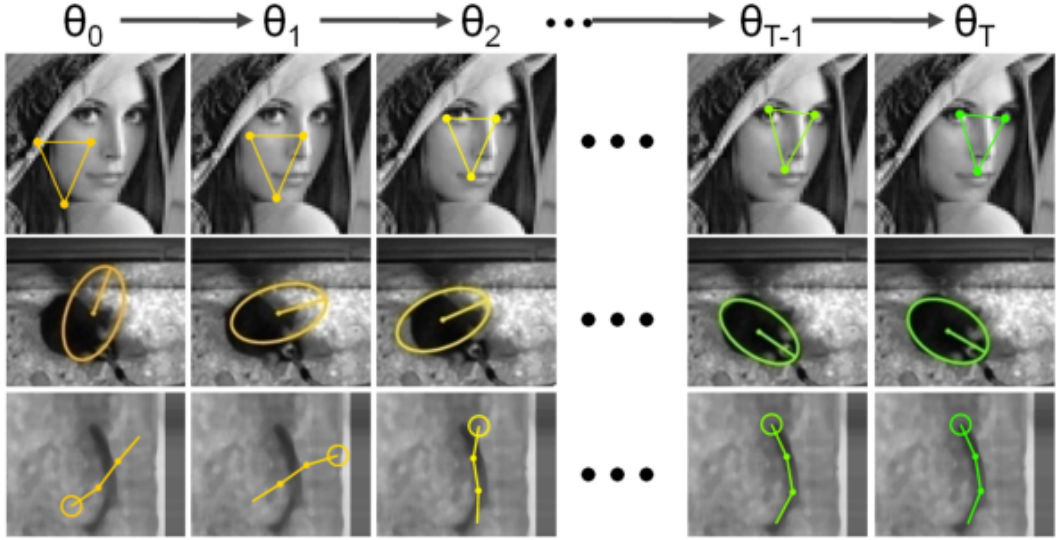}}
\caption{The procedure of refinements on poses in 3 different cases. The researchers generate an initial pose arbitrarily and use a regressor $R^t$ to refine pose $\theta_t$, where $t=1...T$. Regressors $R^t$ are computed in a training procedure}
\end{figure*}
Inspired by random fern classifier and random forests regressor, the researchers train a random fern regressor at each phase in the cascade. A fern regressor outputs $y_i\in\mathbb{R}$ by taking the featurs $x_i\in\mathbb{R}^F$ as input. Given that a Fern is generated by $S$ pairs of random features from all the features in the image and one pair of features can turn to be a value of 1 or 0, each $x_i$ can be viewed as a value range from 0 to $2^S-1$. $y$ is the prediction for the mean of $y_i$'s of the training examples matching the value of each $x_i$.  This method can converge very quickly and accurately in predicting 3D positions of body joints from a single depth map, using no temporal information. CPR may fail to make a correct pose estimate due to some bad initial pose. To improve the performance of CPR, the researchers find the region that have more sucessful initial poses by running CPR $K$ times with a variety of random intial poses for every image.

\section*{Experiments and results}
\begin{figure*}[h]
\centerline{\epsfig{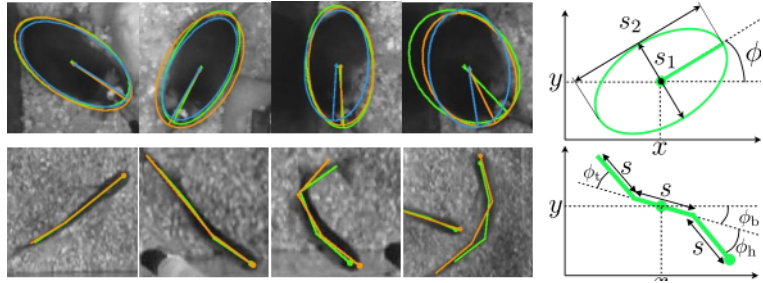}}
\caption{Four left rows images: annotations for Mice and Fish datasets with different color denotes annotator. The rightmost two images: parameterization of the poses. The mose pose is an ellipse at location $(x,y)$ with oreientation $\phi$, scale $s_1$ and aspect ratio $s_1/s_2$. The fish pose is a 3-part model where the body (middle) part is centered at location $(x,y)$ with orientation $\phi_b$, tail part has an angles $\phi_t$, and head part has an angle $\phi_h$}
\end{figure*}
This section covers the experiments and results CPR on two datasets: Mice and Fish. Human annotations are used to make poses of mice and fish clearer, which are shown in Figure 5. For each mouse, its pose include the location, orientation, scale and aspect ratio of an ellipse fitting around For each fish, its pose is a 3-part model with location, orientation and scale of a central body part, and the angles of the tail and head with respect to the body. Figure 6 shows how the number of phases influence the error. The alogrithm converges after 512 stages for both datasets.
\begin{figure*}[h]
\centerline{\epsfig{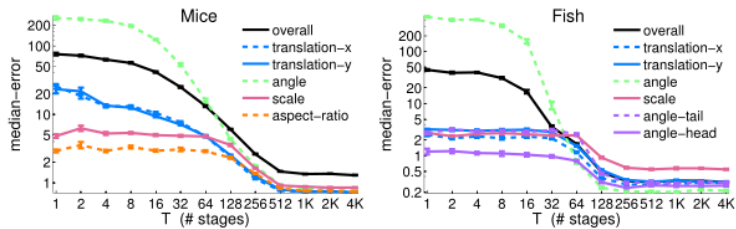}}
\caption{Performance vs. the number of phases T in CPR}
\end{figure*}

\begin{figure*}[!htb]
\centerline{\epsfig{figure=fig6, width=15cm}}
\end{figure*}

\begin{figure*}[!htb]
\centerline{\epsfig{figure=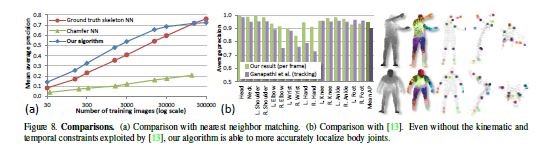, width=15cm}}
\end{figure*}

\end{document}